\documentclass[10pt,twocolumn,letterpaper]{article}

\usepackage{iccv}
\usepackage{times}
\usepackage{epsfig}
\usepackage{graphicx}
\usepackage{amsmath}
\usepackage{amssymb}

\usepackage[breaklinks=true,bookmarks=false]{hyperref}

\iccvfinalcopy

\def\iccvPaperID{????}

\begin{document}

%%%%%%%%% TITLE
\title{Dueling Decoders: Regularizing Variational Autoencoder Latent Spaces}

\author{
Bryan Seybold, Emily Fertig, Alex Alemi, Ian Fischer\\
Google Research\\
{\tt\small \{seybold,emilyaf,alemi,iansf\}@google.com}
}

\maketitle

%%%%%%%%% ABSTRACT
\begin{abstract}
Variational autoencoders learn unsupervised data representations, but these models frequently converge to minima that fail to preserve meaningful semantic information. For example, variational autoencoders with autoregressive decoders often collapse into autodecoders, where they learn to ignore the encoder input. In this work, we demonstrate that adding an auxiliary decoder to regularize the latent space can prevent this collapse, but successful auxiliary decoding tasks are domain dependent. Auxiliary decoders can increase the amount of semantic information encoded in the latent space and visible in the reconstructions. The semantic information in the variational autoencoder's representation is only weakly correlated with its rate, distortion, or evidence lower bound. Compared to other popular strategies that modify the training objective, our regularization of the latent space generally increased the semantic information content.
\end{abstract}

%%%%%%%%% BODY TEXT
\section{Introduction}
For many problem domains, large quantities of data are available without labels. One approach to leveraging such data is to convert the high-dimensional raw data into a low-dimensional \textit{representation} using some unsupervised learning procedure. A desirable property in a learned representation is that nearby representations share semantically meaningful information about the data, such as task labels or world state~\cite{bengio2013representation}. Among the many uses of representation learning are simplifying reinforcement learning (e.g.,~\cite{hafner2018learning}) and utilizing unlabeled data for semi-supervised classification tasks (e.g.\,\cite{kingma2014semi,alex2018therml}).

One popular method for unsupervised representation learning is \textbf{variational autoencoders (VAEs)} ~\cite{kingma2013auto}. VAEs are a probabilistic extension of autoencoders, which are themselves non-linear extensions of principal components analysis ~\cite{kingma2013auto, bengio2013representation}. VAEs learn a generative model of the data via approximate variational inference and reconstruct inputs encoded into a probabilistic latent space. The latent space is a learned representation that often has the property that nearby points share the same task label. One criticism of early VAEs was that the reconstructions and samples they produce are often of lower visual quality compared to models such as generative adversarial networks~\cite{goodfellow2014generative}. The introduction of autoregressive decoders in VAEs led to models that produce outputs of better visual quality but often ignore the latent space~\cite{oord2016pixel, van2016conditional, salimans2017pixelcnn, gulrajani2016pixelvae, chen2016lossy, kingma2016improving, bowman2015generating}. Researchers have modified the VAE training objective in several ways, such as ``free bits'', $KL$ annealing, Penalty-VAE, or adding an auxiliary decoder to encourage models to utilize the latent space~\cite{bowman2015generating, kingma2016improving, higgins2017beta, alemi2018fixing, burgess2018understanding, zhao2017learning}. The theoretical basis of VAEs for representation learning is still being developed and the practical consequences of modifying the training objective for learned representations is not yet understood~\cite{chen2016lossy, burgess2018understanding, alemi2018fixing}.

\begin{figure}
    \centering
    \begin{minipage}[b]{0.95\columnwidth}
    \centering
    \includegraphics[width=\columnwidth]{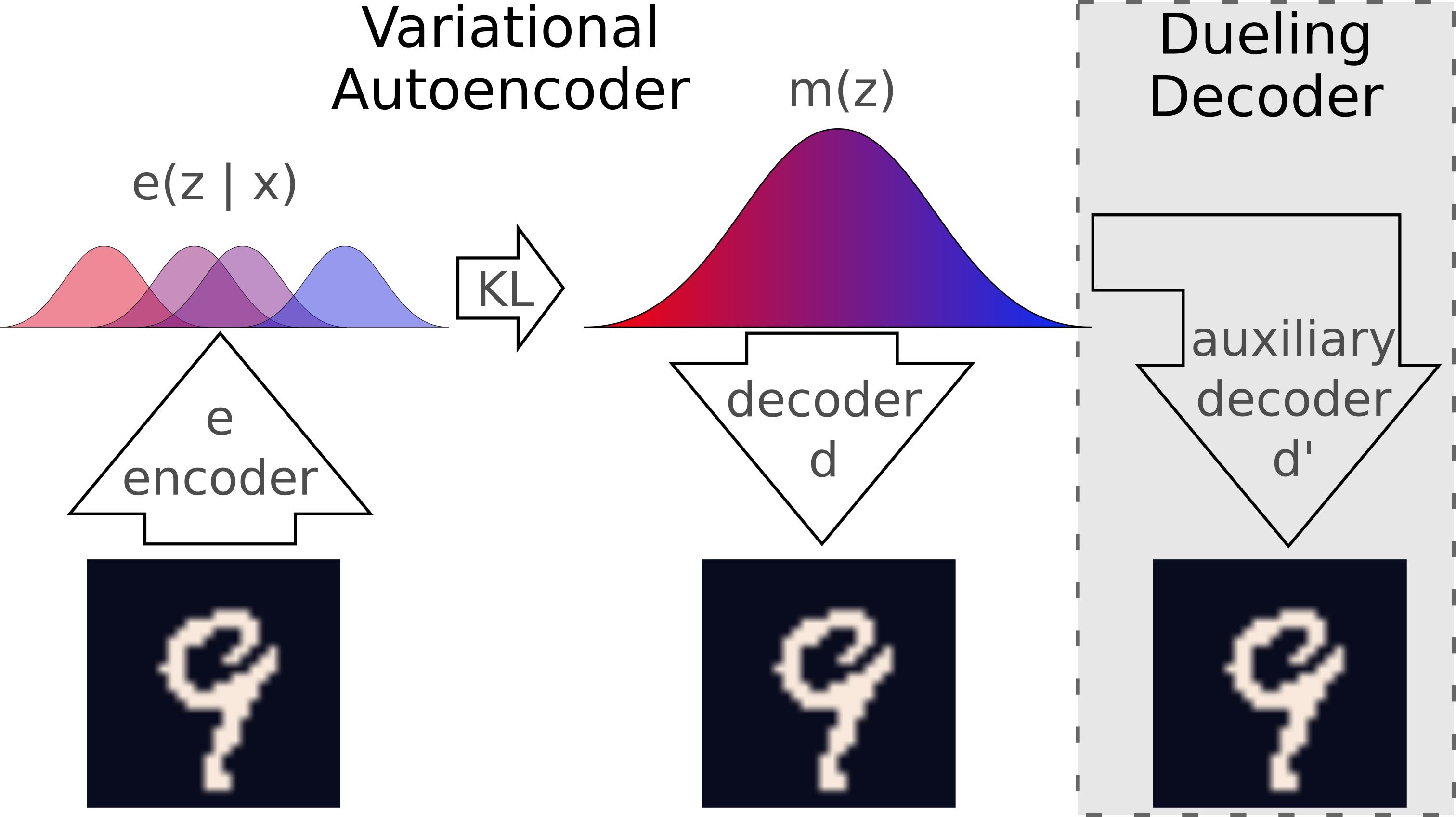}\\
    %\footnotesize{Model Overview}
    \end{minipage}
    \vspace{0.1cm}
    \caption{Overview of a variational autoencoder with auxiliary decoder for regularization. Images are encoded on the left into an encoder distribution, $e(z|x)$. The approximate marginal distribution, $m(z)$, learns to match the mixture of all of the encoder distributions across the training data, and samples from the encoder or the marginal can be decoded by sampling from the decoder $d(x|z)$. The additional, lightly shaded, auxiliary decoder $d'(x|z)$ is a central focus of our paper---a regularizer that helps to learn more semantically meaningful representations.
    }
    \label{fig:model_overview}
    \vspace{-0.5cm}
\end{figure}

In this work, we explore how modifications of learning objective (but not the architecture or optimization procedure) influence the semantic information in the learned representation and reconstructions. We measure the ability of the network to learn a semantically meaningful representation by observing whether the encoded input can be decoded into the semantic label and whether reconstructed images maintain the same label as the inputs. We demonstrate that common measures of VAE performance, including the \textbf{evidence lower bound (ELBO)}, rate, and distortion, may be poor proxies for measuring semantic information retained in the latent representation. We identify that adding an auxiliary decoder can greatly improve the semantic information in the learned representation, but that different auxiliary decoders are required for image labeling than for natural language processing.

To summarize, our main contributions are:
\begin{itemize}%[leftmargin=0]
    \vspace{-1ex}
    \setlength{\itemsep}{0pt}
    \setlength{\parskip}{0pt}
    \setlength{\parsep}{0pt}
    \item a comparison across training objective variants showing that auxiliary tasks improve semantic information efficiently relative to other approaches; 
    \item a demonstration that the effective auxiliary tasks for computer vision tasks can be distinct from effective auxiliary tasks for natural language processing;
    \item a simple procedure for measuring the semantic information maintained in VAE reconstructions that measures how well the decoder represents semantic information;
    \item that the semantic information retained by VAEs is not well predicted by the ELBO, distortion, or rate of the VAE;
    \item that VAEs with the same architecture and rate encode latent spaces with different amounts of semantic information;
    \item that autoregressive decoders suffer from local minima that lead to posterior collapse early in training, but after converging to a good ELBO, models are stable.
\end{itemize}

\section{Related Work}
\subsection{Understanding Autoregressive Decoders Training Objectives}
PixelRNN introduced the first tractable autoregressive artificial neural network generative model of images~\cite{oord2016pixel}. This was extended to feed-forward convolutional networks with PixelCNN and PixelCNN++~\cite{oord2016pixel, van2016conditional,salimans2017pixelcnn}. These generative models can achieve the same or better log probability as VAEs on many image tasks even without conditioning on an input image~\cite{kingma2013auto, alemi2018fixing, oord2016pixel}.
It is natural to consider combining these more powerful generative models with the VAE framework, but the naive approach results in the trained decoder ignoring the latent representation and becomes an auto\underline{de}coder instead~\cite{chen2016lossy, alemi2018fixing, van2017neural}.
\cite{chen2016lossy, alemi2018fixing} claim that any sufficiently powerful decoder will collapse into an autodecoder when optimizing the ELBO objective.
Modifications to the ELBO have been proposed to mitigate this issue.
The simplest such modification, the $\beta$-VAE~\cite{higgins2017beta}, adds a Lagrange multiplier to the $KL$ term of the ELBO.
Other efforts have been made to better understand and control VAE optimization with autoregressive decoders,
including $KL$ annealing~\cite{bowman2015generating}, allowing the model ``free bits''~\cite{kingma2016improving, chen2016lossy}, Penalty-VAE~\cite{alemi2018fixing,burgess2018understanding}, or adding an auxiliary decoding task~\cite{zhao2017learning}.
More drastic changes modified the optimization procedure or modified the model architecture to try to circumvent these issues~\cite{Zhao2017, yang2017improved, kim2018semi, dieng2018avoiding, he2019lagging}. Changing the optimization procedure or architecture is in some senses orthogonal to changes in the loss, so we limit our study to understanding how the loss shapes the learned representation.
In this work, we compare many modifications to the ELBO using classification-based measures of semantic information in the learned representations.

\subsection{Quantifying Learned Representations}
A common method of estimating the quality of the learned latent representation is through visual inspection~\cite{kingma2016improving} or reporting numbers that are based on the training objective, such as bits-per-pixel or rate.
In contrast, we seek to measure the amount of information about image labels in the encoded space and reconstructions. While the rate is the upper-bound on the semantic information, the information regarding labels or other semantics can be much lower.
One means of assessing learned representations is to assess ``disentanglement'' where semantically meaningful information falls along independent axes or manifold directions in the latent space~\cite{higgins2017beta, burgess2018understanding}. Both~\cite{higgins2017beta}~and~\cite{burgess2018understanding} increase disentanglement by increasing $\beta$, which can exacerbate posterior collapse with PixelCNN models~\cite{alemi2018fixing}.
It is unclear how discrete labels should map onto a small number of disentangled axes, so we do not evaluate disentanglement.
An alternative method of quantifying learned representation quality is measuring the classification accuracy from latent space values to semantic labels. Linear classifiers are associated with interpretability, but work in low dimensional latent spaces shows intuitively that representations do not need to be linearly separable in order to be easily interpretable~\cite{bengio2013representation,alemi2016vib,alemi2018uncertainty,davidson2018hyperspherical,anonymous2019the,fertig2018beta}. None of these methods examine the end-to-end quality of the VAE including whether the reconstructed images respect semantic information in the input.

\subsection{Regularization and Auxiliary Losses}
Many approaches to ease training and improve generalization have been proposed for deep networks.
Some of the most popular are BatchNorm and other forms of normalization~\cite{ioffe2015batch,ba2016layer,wu2018group}, Dropout~\cite{srivastava2014dropout}, adding skip-connections~\cite{he2016deep, dieng2018avoiding}, and L1 or L2 regularization.
Auxiliary losses are commonly used when training deep networks in a variety of situations, including deterministic classification and object detection~\cite{szegedy2015going}, reinforcement learning~\cite{jaderberg2016reinforcement}, and probabilistic models~\cite{vedantam2017generative}. Recent work in probabilistic models has explored a variety of what may be considered auxiliary losses in the VAE family of objectives.
\cite{zhao2017learning} proposed a bag-of-words auxiliary decoder to a variational autoencoder for natural language processing.
The unpublished \cite{lucas2018auxiliary} proposed adding a similar pixel-independent auxiliary loss to a PixelVAE architecture.
Neither~\cite{zhao2017learning}~nor~\cite{lucas2018auxiliary} measured the semantic information extracted by the autoencoder. 
\cite{Spurr_2018_CVPR} shares the same latent space across multiple encoders and decoders for cross-modal learning.
\cite{chen2016lossy} discusses limiting autoregressive models to only generated portions of an image.
\cite{Zhao2017} adds an adversarial discriminator network to prevent the VAE from becoming an autodecoder.
\cite{kingma2014semi} and \cite{alex2018therml} added a classifier on the latent decoder to perform semi-supervised learning.
We are unaware of any work that attempts to quantify the quality of the resulting learned representations, which is the core focus of this work.

\section{Regularizing Latent Spaces with Dueling Decoders}
\subsection{Preliminaries}
Variational inference is an approximate Bayesian method for fitting the parameters of generative models ~\cite{Bishop:2006:PRM:1162264, Murphy:2012:MLP:2380985}.
Following the notation of~\cite{alemi2018fixing}, given data $X$, we may learn a representation $Z \leftarrow X$ by selecting a parameterized \textit{encoder} $e(z|x)$, a parameterized \textit{marginal} $m(z)$, and a parameterized \textit{decoder} $d(x|z)$, and maximizing the ELBO~\cite{kingma2013auto}:
\begin{align}\label{eq:elbo_kl}
\mathbb{E}_{e(z|x)}[\log d(x|z)] - D_{KL}[e(z|x)||m(z)]
\end{align}
where $D_{KL}$ is the Kullbach-Leibler Divergence.
This objective may be augmented with a hyperparameter, $\beta$~\cite{higgins2017beta}:
\begin{align}\label{eq:elbo_b_kl}
\mathbb{E}_{e(z|x)}[\log d(x|z)] - \beta D_{KL}[e(z|x)||m(z)]
\end{align}
This hyperparameter may be viewed as a regularization that encourages disentangled representations $Z$~\cite{higgins2017beta, burgess2018understanding}, or from an information-theoretic perspective where $\beta$ controls the \textit{rate} ($R$) of information $Z$ maintains about $X$~\cite{alemi2018fixing}:
\begin{align}
R \equiv \int dx\, p(x) \int dz\, e(z|x) \log \frac{e(z|x)}{m(z)} \ge I(X;Z)
\end{align}
where $I(X;Z)$ is the mutual information between $X$ and $Z$. $R$ is the $D_{KL}$ term from the ELBO and bounds the information extracted from the input.
Similarly, we may identify the \textit{distortion} ($D$) as the first term of the ELBO:
\begin{align}
D &\equiv \int dx\, p(x) \int dz\, e(z|x) \log d(x|z) \nonumber \\
&\le I(X;Z) - H(X)
\end{align}
Thus, the ELBO may be understood as simultaneously minimizing an upper bound and maximizing a lower bound on $I(X;Z)$, and $\beta$ as a Lagrange multiplier that allows us to trade off between those two bounds.

\subsection{Semantic Representation Accuracy}
\label{sec:sem_latent}
While the rate is an upper bound on information extracted from the VAE inputs, it does not require that semantically relevant information is extracted. We may directly measure the semantic information in the latent representation by separately training a classifier $P(Y|Z)$ on a labeled training set $X,Y$ and corresponding $Z$ sampled $N$ times from $e(z|x)$ for each $x \in X$ and measuring the accuracy of the classifier on a held-out labeled test set:
\newcommand{\argmax}{\mathop{\mathrm{argmax}}}
\begin{align}\label{eq:latent_score}
\frac{1}{N}\sum_{N} \text{\bf 1}_{Y = \argmax P(Y|Z)},
\end{align}
where $\text{\bf 1}_{p}$ is the indicator function (1 when $p$ is true, 0 otherwise). The label distortion is an information theoretic measurement of the semantic information in the latent encoding derived as the classifier's conditional entropy:
\begin{align}\label{eq:label_distortion}
H(Y|Z) = H(Y) - I(Y;Z).
\end{align}

\subsection{Semantic Reconstruction Accuracy}
\label{sec:sem_recon}
To quantify semantic information from end-to-end of the VAE, we propose a straightforward measurement of the accuracy with which reconstructions maintain the original class label of a dataset. For a dataset, $X$, with labels, $Y$, and $N$ reconstructions of the data from a VAE, $X'$, we train a classifier to maximize the likelihood of $P(Y|X)$ on the original training set, $X$, and determine the \textbf{semantic reconstruction accuracy} of the model as 
\begin{align}\label{eq:class_score}
\frac{1}{N}\sum_{N} \text{\bf 1}_{Y = \argmax P(Y|X')}
\end{align}
on the reconstructions from the test set, $X'$.
In other words, we train a classifier and then measure the classification accuracy on reconstructions $x' \sim p(x) e(z|x) d(x'|z)$.

\subsection{Dueling Decoders}
\label{sec:dueling_decoders}
The typical VAE training objective has no term to explicitly maximize semantic information. The semantic labels needed for such a term are not available during unsupervised learning. The only option in unsupervised learning is modifying the optimization to encourage semantic information. One strategy is to find auxiliary tasks that correlate with semantic information. Both~\cite{zhao2017learning}~and~\cite{lucas2018auxiliary} added auxiliary tasks to VAEs, but only to prevent rate collapse. Neither studied which auxiliary tasks are appropriate for computer vision or measured how the auxiliary task changed semantic information extracted by the autoencoder.

We formalize auxiliary tasks as adding an auxiliary decoder, ${d'}(x|z)$. This decoder shares the same marginal distribution and latent space as the original decoder. Because two different decoders are competing to shape the latent space, we call this method \textbf{Dueling Decoders}. The auxiliary decoder adds another reconstruction term to the training objective.
\begin{align}
\mathbb{E}_{e(z|x)}[\log d(x|z) + \lambda \log d'(x|z)] - \beta D_{KL}[e(z|x) || m(z)]
\label{eq:dueling_decoder}
\end{align}

It is worth noting that this regularization does not change the probabilistic interpretation of the distortion, rate, or ELBO of the original model. During evaluation, the auxiliary decoder is removed and the model is evaluated as a standard VAE. The auxiliary decoder only influences what is learned in the latent space of the model.  This makes it distinct from methods that change the optimization or architecture~\cite{Zhao2017, yang2017improved, kim2018semi, dieng2018avoiding, he2019lagging}.

This formalization holds across a wide range of auxiliary decoding tasks, but does not reveal the optimal auxiliary task to increase semantic information. The search for optimal auxiliary tasks is a primary pursuit in unsupervised learning. For example, if $d(x|z) \propto d'(x|z)$, Eq.~\ref{eq:dueling_decoder} is just reweighting the the terms in the original ELBO. In contrast, if the primary decoder is a PixelCNN and the auxiliary decoder reconstructs the color histogram, the latent space feeding into the PixelCNN will be encouraged to encode additional color information. If this color information is related to label information, it will ensure that semantic information related to the labels is available. Alternatively, if the primary decoder is a PixelCNN and the auxiliary decoder is a pixel-independent CNN decoder, the model will be encouraged to learn global structure. The ideal auxiliary decoder will depend on the exact downstream task and model architecture.

Because the ELBO is bounded by the information in the data set, adding an auxiliary decoder can at best change where the model settles in the rate-distortion plane. Adding regularization breaks the symmetry of points within the ELBO objective similarly to how manipulating $\beta$ changes the rate-distortion performance but can have little effect on the final ELBO~\cite{alemi2018fixing}. However, an outstanding question is whether regularizing the latent space with an auxiliary decoder captures different amounts of semantic information than modifying the ELBO objective terms.

\begin{figure*}
    \centering
    \includegraphics[width=\textwidth]{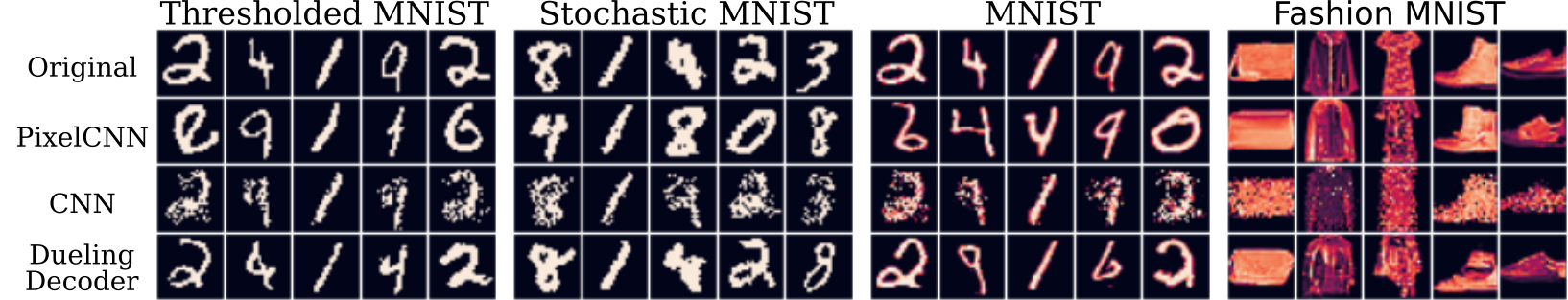}\\
    \vspace{0.1cm}
    \caption{Reconstructions from each decoder class with rate$<$10 and the best semantic reconstruction accuracy. The top row is the original image from the data set. The rows beneath are reconstructions from the PixelCNN, CNN, and Dueling Decoder decoders respectively.}
    \label{fig:reconstructions}
\end{figure*}

\begin{table*}%[b!h]
\centering
\setlength\tabcolsep{5pt}
	\begin{tabular}{|c|c|c|c|c|c|c|c|c|}
	\hline
Decoder & $\beta$ & ELBO & Distortion & Rate & \shortstack{Latent\\Accuracy\\MLP} & \shortstack{Label\\Distortion\\MLP} & \shortstack{Latent\\Accuracy\\Linear} & \shortstack{Reconstruction\\Accuracy} \\
\hline
\multicolumn{9}{|c|}{\textbf{STOCHASTIC MNIST}} \\
\hline
PixelCNN & 0.1 & 83.2 & 74.6 & 8.57 $\pm$ 0.16 & 0.48 $\pm$ 0.03 & 1.34 $\pm$ 0.08 &  0.36 $\pm$ 0.05  & 0.36 $\pm$ 0.03  \\
\hline
CNN & 1.0 & 136.7 & 129.8 & 6.91 $\pm$ 0.09 & 0.81 $\pm$ 0.03 & 0.65 $\pm$ 0.04 &  \textbf{0.55 $\pm$ 0.12}  & \textbf{0.73 $\pm$ 0.02}  \\
\hline
Dueling Decoder & 1.0 & 81.3 & 72.1 & 9.25 $\pm$ 0.72 & \textbf{0.91 $\pm$ 0.02} & \textbf{0.31 $\pm$ 0.06} &  \textbf{0.72 $\pm$ 0.06}  & \textbf{0.79 $\pm$ 0.03}  \\
\hline
\multicolumn{9}{|c|}{\textbf{THRESHOLDED MNIST}} \\
\hline
PixelCNN & 0.1 & 60.6 & 52.6 & 8.01 $\pm$ 0.18 & 0.58 $\pm$ 0.06 & 1.07 $\pm$ 0.14 &  0.39 $\pm$ 0.02  & 0.47 $\pm$ 0.07  \\
\hline
CNN & 1.0 & 130.4 & 123.4  & 7.04 $\pm$ 0.11 & 0.82 $\pm$ 0.01 & 0.59 $\pm$ 0.02 &  \textbf{0.62 $\pm$ 0.07}  & 0.81 $\pm$ 0.02  \\
\hline
Dueling Decoder & 1.0 & 56.4 & 48.4 & 7.93 $\pm$ 0.18 & \textbf{0.92 $\pm$ 0.01} & \textbf{0.27 $\pm$ 0.02} &  \textbf{0.70 $\pm$ 0.02}  & \textbf{0.88 $\pm$ 0.01}  \\
\hline
\multicolumn{9}{|c|}{\textbf{MNIST}} \\
\hline
PixelCNN & 0.1 & 560.1 & 552.8 & 7.29 $\pm$ 0.08 & 0.42 $\pm$ 0.04 & 1.52 $\pm$ 0.06 &  0.27 $\pm$ 0.03  & 0.37 $\pm$ 0.02  \\
\hline
CNN & 1.0 & 891.6 & 883.6 & 7.94 $\pm$ 0.28 & \textbf{0.79 $\pm$ 0.01} & \textbf{0.58 $\pm$ 0.03} &  \textbf{0.51 $\pm$ 0.06}  & \textbf{0.78 $\pm$ 0.01}  \\
\hline
Dueling Decoder & 1.0 & 564.6 & 554.8 & 9.78 $\pm$ 0.96 & \textbf{0.88 $\pm$ 0.04} & \textbf{0.41 $\pm$ 0.11} &  \textbf{0.58 $\pm$ 0.05}  & \textbf{0.79 $\pm$ 0.04}  \\
\hline
\multicolumn{9}{|c|}{\textbf{FASHION MNIST}} \\
\hline
PixelCNN & 1.0 & 1492.2 & 1484.0 & 8.24 $\pm$ 0.17 & \textbf{0.78 $\pm$ 0.01} & \textbf{0.63 $\pm$ 0.03} &  \textbf{0.52 $\pm$ 0.02}  & \textbf{0.76 $\pm$ 0.02}  \\
\hline
CNN & 1.0 & 1995.3 & 1986.0 & 9.31 $\pm$ 0.04 & 0.71 $\pm$ 0.01 & 0.82 $\pm$ 0.02 &  \textbf{0.54 $\pm$ 0.03}  & 0.56 $\pm$ 0.01  \\
\hline
Dueling Decoder & 1.0 & 1495.1 & 1486.4 & 8.76 $\pm$ 1.34 & \textbf{0.72 $\pm$ 0.05} & \textbf{0.79 $\pm$ 0.17} &  \textbf{0.47 $\pm$ 0.08}  & \textbf{0.69 $\pm$ 0.03}  \\
\hline
\hline
PixelCNN+ & 1.0 & 1473.3 & 1469.2 & 4.08 $\pm$ 0.81 & \textbf{0.49 $\pm$ 0.14} & \textbf{1.23 $\pm$ 0.28} &  \textbf{0.30 $\pm$ 0.05}  & 0.43 $\pm$ 0.12  \\
\hline
Dueling Decoder+ & 1.0 & 1480.2 & 1471.2 & 9.03 $\pm$ 0.11 & \textbf{0.68 $\pm$ 0.02} & \textbf{0.94 $\pm$ 0.05} &  \textbf{0.36 $\pm$ 0.06}  & \textbf{0.67 $\pm$ 0.01}  \\
\hline
	\end{tabular}
\vspace{0.2cm}
\caption{For each data set and each decoder, we report results for the model with rate$<$10 and the highest semantic representation accuracy (MLP). Values are the means of three randomly initialized training runs $\pm$ one standard deviation. Bold values are not significantly different from the best value (t-test, equal variance, $\alpha$=0.05, Bonferroni corrected n=2). Decoders marked with + are enlarged networks.}
\label{table:best_results}
\end{table*}

\section{Experiments}
\subsection{Implementation Details}
We evaluate VAEs with different decoder architectures and hyperparameters on 4 data sets. Two data sets are binary versions of MNIST~\cite{lecun2010mnist}: Stochastic MNIST ~\cite{salakhutdinov2008quantitative, alemi2018tweet} and Thesholded MNIST ~\cite{fischer2014training}. In Stochastic MNIST, pixel values were assigned to 0 or 1 with a probability dependent on their intensity. In Thresholded MNIST, all pixel values are assigned to 1 if greater than 127 and to 0 otherwise. Stochastic MNIST is commonly used in the VAE literature ~\cite{kingma2013auto, alemi2018fixing, kingma2016improving, kingma2014semi, tomczak2017vae}, but the process of sampling pixel values independently decreases the degree of local correlation in the images. Thresholded MNIST preserves local correlations in each image. We also use two 8-bit data sets, MNIST ~\cite{lecun2010mnist} and Fashion MNIST ~\cite{xiao2017fashion}. Each data set has 10 labels and label entropy, $H(Y)$, of ~2.3 nats.

For each data set, we train a range of VAEs. (The full architectures are listed in the supplemental). All VAEs use the same convolutional neural network architecture for the encoder and a 280 component VAMPPrior marginal ~\cite{tomczak2017vae}. The encoder outputs are means and lower triangular representations of covariance matrices of multivariate normal distributions. 
We use three different types of decoders: pixel-independent convolution decoders (CNN), autoregressive PixelCNN++ decoders ~\cite{salimans2017pixelcnn} (PixelCNN), PixelCNN++ decoders with auxiliary decoders (Dueling Decoders) as described in~Section~\ref{sec:dueling_decoders}. Unless otherwise noted, the auxiliary decoder is a pixel-independent convolutional decoder.
% that the PixelVAE decoder has as many parameters as the CNN and PixelCNN models combined, and
Note that, at test time, the PixelCNN and Dueling Decoder architectures are identical. The architectures were chosen to achieve ELBOs between 80 and 90 nats on Stochastic MNIST (approximately matching results in the literature~\cite{alemi2018fixing, kingma2016improving}), and we fixed the architecture to see if the model could generalize across data sets. For binary data sets, the decoder output distribution is a Bernoulli, and for 8-bit data sets, the output distribution is a mixture of quantized logistics~\cite{salimans2017pixelcnn}.
We perform a grid search over the following parameters for each model: $\beta$ (0.1, 1.0), training batch size (32, 64), number of latent dimensions (2, 16, 64), and dueling decoder regularization $\lambda$ (0.1, 1.0). We train each model for 20000 steps (12 or 25 epochs depending on batch size) with the Adam optimizer and time varying learning rate,
$1^{-3} * 0.66^{t / 1^{-4}} * (1 - 0.66^{ t / 1^{-3}}) + 1^{-10}$, where $t$ is the step number.
Reconstructions from different models with rate$<$10 are available in Figure~\ref{fig:reconstructions}. 

To measure the semantic representation accuracy, we trained a linear classifier and a multilayer perceptron (MLP) from the encoded latent values from the test set and report the accuracy of those classifiers on the test set as described in~Section~\ref{sec:sem_latent}.

To measure the semantic reconstruction accuracy, we train a convolutional neural network classifier on each data set. The network is trained on the original training set and validated on the test set to have $>$98\% accuracy on each MNIST variant and 94\% accuracy on the more difficult Fashion MNIST data set. The classifier is only trained on the original training set, not on reconstructions. The classifier was applied to 300 evenly sampled reconstructions from the test set to test how well the reconstructions match the semantics of the original data set as described in~Section~\ref{sec:sem_recon}.

\subsection{Dueling Decoders exceed the best of independent and autoregressive decoders.}

\begin{figure}
    \centering
    \begin{minipage}[b]{0.9\columnwidth}
    \includegraphics[width=\columnwidth]{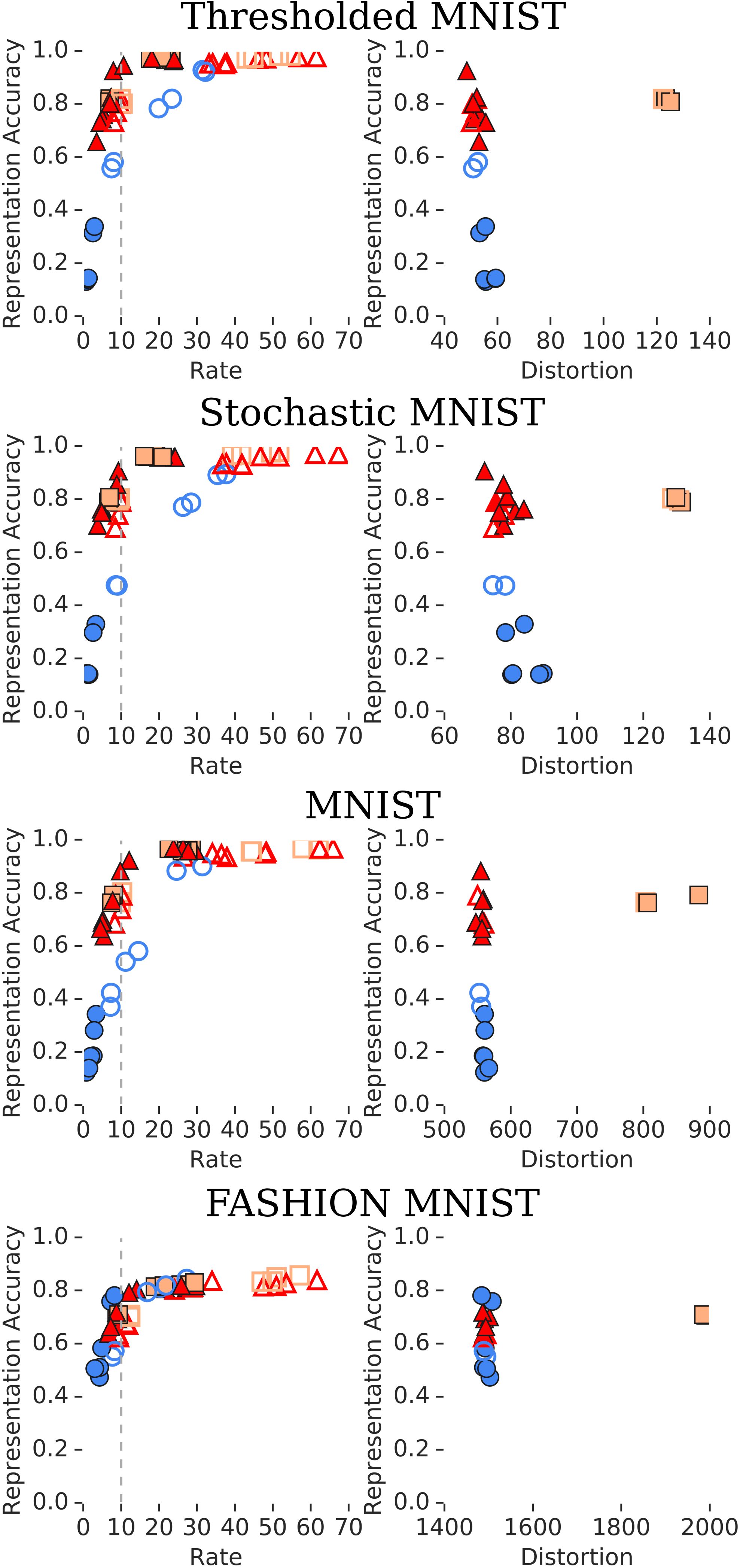}\\
    \end{minipage}
    \vspace{-0.2cm}
    \caption{Correlations between semantic representation accuracy and the rate or distortion. \textbf{left:} Rate versus accuracy for models marked by decoder class: 
    PixelCNN (\textcolor{blue}{$\bullet$}), 
    CNN (\textcolor[rgb]{1.0, 0.5, 0.0}{$\blacksquare$}), and 
    Dueling Decoder (\textcolor{red}{$\blacktriangle$}). Filled or empty markers denote $\beta=1.0$ or $\beta=0.1$ respectively. The vertical dashed line indicates the 10 nat threshold for low-rate. \textbf{right:} Distortion versus accuracy for the low-rate models. Best viewed in color.}
    \label{fig:accuracy_correlations}
    \vspace{-0.5cm}
\end{figure}

Although decoder architectures are not the focus of this paper, it is illustrative to show how the choice of decoder changes the semantic information the VAE learns to extract (Figure~\ref{fig:reconstructions} and Table~\ref{table:best_results}). In agreement with~\cite{alemi2018fixing}, models with the same decoder demonstrate a trade off in rate-distortion space due to the limited amount of information in the data set (see supplement). We focus further analysis on these low-rate models because the rate versus accuracy curve in Figure~\ref{fig:accuracy_correlations} has an inflection point just above 10 indicating that accuracy increases most rapidly at low rates, the entropy of all four training sets is upper-bounded by $\log 60000 = 11~\text{nats}$ (the number of data values), and compact representations are a desirable property. Results across the range of low rate models ($<$20 nats) show similar trends (Fig.~\ref{fig:accuracy_correlations}). Across Thresholded, Stochastic, and the original MNIST, VAEs with autogregressive PixelCNN decoders achieved low distortions, but encoded little semantic information (all semantic accuracies $<$ 0.6). In contrast, VAEs with independent CNN decoders encoded more semantic information (reconstruction and MLP semantic accuracies $>$ 0.7), but suffered from high distortion. (At higher rates, CNN decoders can also achieve good distortions on MNIST, but not at rates less than 10 nats.) 
In contrast, using a PixelCNN architecture with an auxiliary CNN decoder in a Dueling Decoder setup achieves both low distortions and high semantic information (reconstruction and MLP accuracies $>$ 0.7). Notably, the Dueling Decoder strategy achieves better distortions, rates, and trends towards better semantic accuracies on MNIST than either the PixelCNN or CNN decoders alone.

It worth noting that the PixelCNN models with the highest representation accuracy on MNIST variants were trained with $\beta=0.1$ because the rate term collapses to near zero for $\beta=1.0$ (Figure~\ref{fig:accuracy_correlations}). Because the rate is an upper bound on any information, models with zero rate cannot contain semantic information. Decreasing $\beta$ or another manipulation is necessary to potentially increase semantic information. In contrast, both CNN decoders and Dueling Decoders maintained a rate larger than the entropy in the labels without manipulating $\beta$. The semantic information retained by CNN and Dueling Decoders is higher than PixelCNN with $\beta=0.1$ even at similar rates. This indicates that the increased rate achieved by decreasing $\beta$ in these models encodes style information or noise rather than semantic information. Decreasing $\beta$ does not preferentially guide the model to encode semantic information.

On Fashion MNIST, PixelCNN at $\beta=1.0$ does not collapse with all hyperparameter combinations (particularly for larger batch sizes). When the model does not collapse, it encodes relevant semantic information well. However, adding a Dueling Decoder consistently prevented collapse and increased semantic information. Adding a Dueling Decoder can therefore simplify hyperparameter searches without hurting performance if unnecessary. Furthermore, requiring a larger batch size to avoid collapse limits the memory available for the network architecture. Increasing the number of layers and number of kernels in the PixelCNN decoder can yield better ELBOs and distortions (1469.2 $\pm$ 3.2 versus 1484.0 $\pm$ 1.7), but limits the batch size to 32. The model collapses again. In contrast, the enlarged Dueling Decoder model trends towards maintaining more information in the latents and generates significantly more semantically accurate reconstructions.

\subsection{Comparing different modified training objectives.}
\label{sec:modification_comparisons}

\begin{table}%[b!h]
\centering
\setlength\tabcolsep{5pt}
	\begin{tabular}{|c|c|c|c|}
\hline
Modification & Rate &  \shortstack{Latent\\Accuracy\\MLP} & \shortstack{Reconstr.\\Accuracy} \\
\hline
PixelCNN $\beta=1$ & 0.97 & 0.14 $\pm$ 0.01 & 0.11 $\pm$ 0.01 \\
\hline
PixelCNN $\beta=0.1$ & 31.67 & 0.93 $\pm$ 0.01 & 0.91 $\pm$ 0.01 \\
\hline
\hline
Dueling Decoder & 8.13 & \textbf{0.92 $\pm$ 0.01} & \textbf{0.88 $\pm$ 0.01} \\
\hline
$KL$ Annealing & 9.16 & \textbf{0.92 $\pm$ 0.01} & \textbf{0.89 $\pm$ 0.02} \\
\hline
Free Bits & 10.16 & 0.81 $\pm$ 0.01 & 0.74 $\pm$ 0.02 \\
\hline
Penalty-VAE & 10.04 & 0.53 $\pm$ 0.05 & 0.41 $\pm$ 0.05 \\
\hline
	\end{tabular}
\vspace{0.1cm}
\caption{Comparison of training modifications to avoid collapsing into an autodecoder on Thresholded MNIST. Bold font indicates values not significantly different from the best value in the bottom four rows (t-test, equal variance, $\alpha$=0.05, Bonferroni corrected n=3).}
\label{table:modification_comparisions}
    \vspace{-0.5cm}
\end{table}	

Because the Dueling Decoder and the PixelCNN have the same architecture at test time, the difference between the two models is only in the objective that they optimize. We compared our modification of the training objective to other modifications of the training objective from the literature in Table.~\ref{table:modification_comparisions}. All models used the same architecture and procedure at inference time. Manipulating $\beta$ resulted in either a near-zero rate and low-accuracy reconstructions or a very-high rate. Dueling Decoders and $KL$ Annealing~\cite{bowman2015generating} both outperformed Free Bits~\cite{kingma2016improving} and Penalty-VAE~\cite{burgess2018understanding, alemi2018fixing} approaches (0.92 and 0.92 semantic representation accuracy vs 0.81 and 0.53). Dueling Decoders achieved similar accuracy at a marginally lower rate and was easier to train because the objective does not change over time as in $KL$ Annealing. As $KL$ annealing and Dueling Decoders modify different portions of the loss, these two could be combined in the future.

\subsection{Auxiliary decoders requiring spatial information increase semantic information.}

All Dueling Decoder results up to this point in the paper have used a pixel-independent CNN auxiliary decoder, but it is initially unclear what makes this auxiliary task effective at increasing semantic information. Previously published work has used a bag-of-words decoder in a natural language processing model~\cite{zhao2017learning}, which corresponds to a color or intensity histogram decoder. To determine how the choice of auxiliary tasks determines the semantic information learned, we tested four decoders (Table~\ref{table:aux_decoders}): 1) an intensity histogram decoder that reconstructing the distribution of intensity values across the entire image, 2) a marginal histogram decoder that reconstructs the distribution of intensity values in each row and each column independently, 3) the CNN decoder that reconstructs the distribution of intensity values in each pixel independently, and 4) an image gradient decoder that reconstructs the distribution of local intensity changes in the vertical and horizontal directions. Later decoders require encoding additional spatial information. Convolutional and image gradient decoders, which require more spatial information, increase the amount of semantic information learned by the VAE over marginal and histogram decoders (latent accuracy - MLP: 0.94 and 0.91 vs 0.82 and 0.50). Notably, an auxiliary histogram decoder does not increase the semantic information relative to the PixelCNN baseline (latent accuracy - MLP: 0.58 vs 0.50). Training supervised classifiers on the reconstruction targets (e.g. intensity histogram) yields the same order of performance. Therefore, it is important to identify the correct auxiliary decoder for the semantics of interest, supervised performance on reconstruction targets is a useful proxy for effectiveness, and natural language processing and image classification require different auxiliary tasks.  

\begin{table}[h]
\small
\centering
\setlength\tabcolsep{5pt}
	\begin{tabular}{|c|c|c|c|c|}
    \hline
    Auxiliary & Rate & \shortstack{Latent\\Accuracy\\MLP}  & \shortstack{Reconst.\\Accuracy} & \shortstack{Supervised\\Accuracy\\on Target} \\
    \hline
    % \multicolumn{5}{|c|}{\textbf{THRESHOLDED MNIST}} \\
    % \hline
None & 7.99 & 0.58 $\pm$ 0.06 & 0.41 $\pm$ 0.05 & \\
\hline
Conv & 9.75 & \textbf{0.94 $\pm$ 0.01} & \textbf{0.90 $\pm$ 0.03} & 0.99\\
\hline
Gradient & 9.38 & 0.91 $\pm$ 0.01 & \textbf{0.88 $\pm$ 0.01} & 0.99\\
\hline
Marginals & 9.03 & 0.82 $\pm$ 0.02 & 0.76 $\pm$ 0.02 & 0.90 \\
\hline
Histogram & 8.09 & 0.50 $\pm$ 0.01 & 0.42 $\pm$ 0.01 & 0.32 \\
    \hline
	\end{tabular}
\caption{Best semantic accuracies at rates less than 10 nats for Thresholded MNIST for each auxiliary decoder and the performance of supervised classifiers trained with each auxiliary target as input. Other data sets omitted for brevity but are similar.}
\label{table:aux_decoders}
\vspace{-0.5cm}
\end{table}

\subsection{The ELBO defines a Valley of Stable Points}

\begin{figure}[b]
    \centering
    \begin{minipage}[b]{0.99\columnwidth}
    \centering
    \includegraphics[width=\columnwidth]{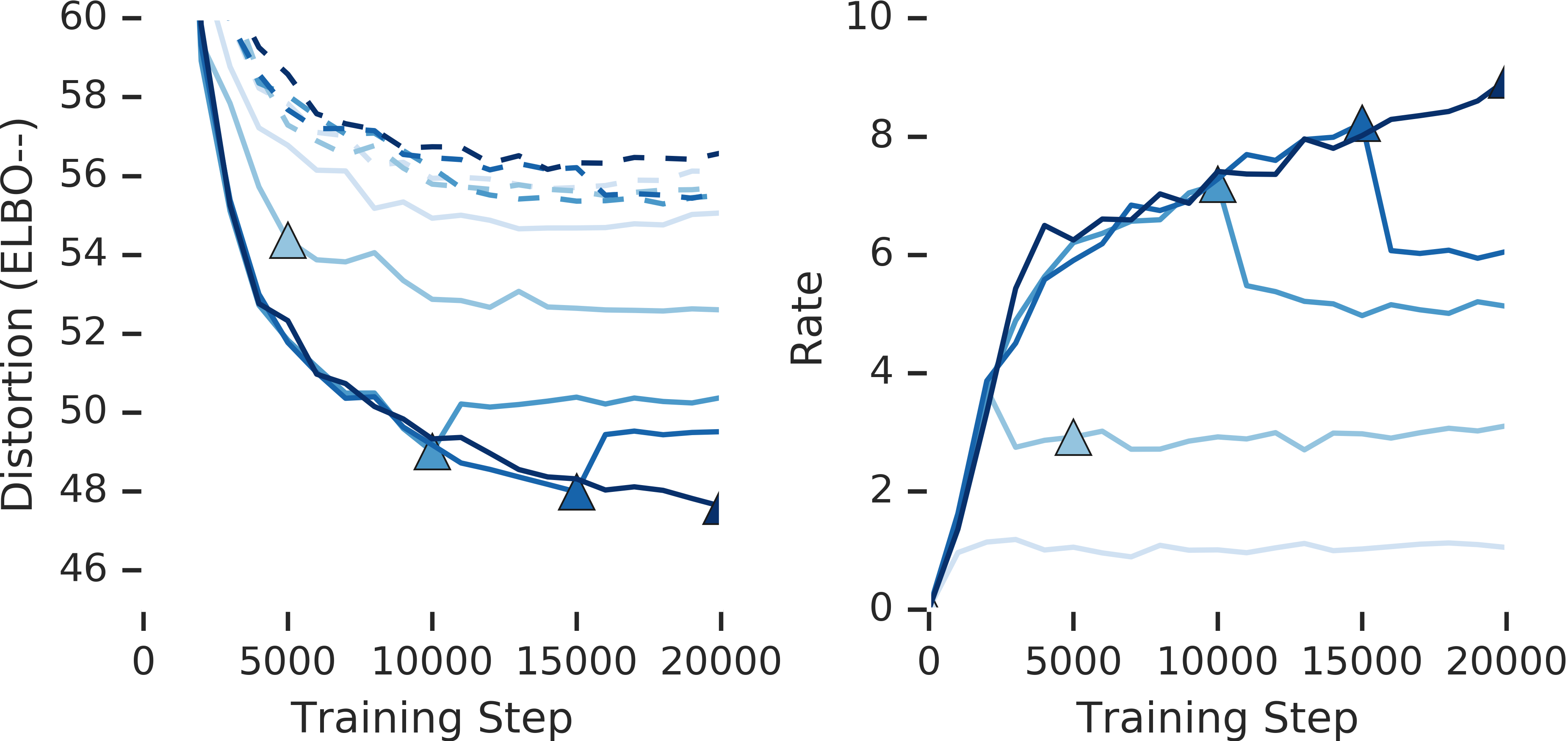}\\
    \end{minipage}
    \vspace{0.1cm}
    \caption{Removing regularization during training produces different convergence points. \textbf{Left:} Distortion (solid lines) and ELBO (dashed lines) when removing regularization at 0, 5000, 10000, or 20000 steps (marked by a $\blacktriangle$ in distortion curve). \textbf{Right:} Similar curve for the rate.}
    \label{fig:dropping_regularization}
\end{figure}

Dueling Decoder models demonstrate much better semantic reconstruction performance on MNIST tasks than purely autoregressive models even at the same rate, distortion, and model architecture. This indicates autoregressive decoders are capable of utilizing semantic information present in the latent space at $\beta=1.0$. However, autoregressive models all converge to similar ELBO values (approximately 56 nats on Thresholded MNIST) even when the training objective is modified to change the rate-distortion trade-off~\cite{alemi2018fixing}. It is not immediately obvious whether all points in parameter space with the same ELBO are stable or whether the structure of PixelCNN decoders encourages collapsing into an autodecoder. Stated another way, do all points with the same ELBO form a valley in optimization space or are autoregressive models biased towards autodecoders? To test this, we took Dueling Decoder models at various points during training, disabled the regularization term, and observed the ELBO, rate, and distortion the models converged to. (The converse is less interesting because adding regularization will necessarily pull the model toward a new point in parameter space.) When removing regularization from models during training, we discovered that the models continued to leverage the latent spacce, but each model converged to a different rate-distortion point with similar ELBOs (Figure~\ref{fig:dropping_regularization}). This is analogous to $KL$-annealing to $\beta=1.0$ performing well~\cite{bowman2015generating, chen2016lossy}. (Under $KL$-annealing, $\beta$ is slowly increased over time, but Dueling Decoders does not require modifying the objective over time.) These results suggests that models at many points with the same ELBO are stable regardless of where they lie in the rate-distortion plane.

\subsection{Latent Structure Explains Semantic Reconstruction Accuracy}

\begin{figure}
    \centering
    \begin{minipage}[b]{0.95\columnwidth}
    \centering
    \includegraphics[width=\columnwidth]{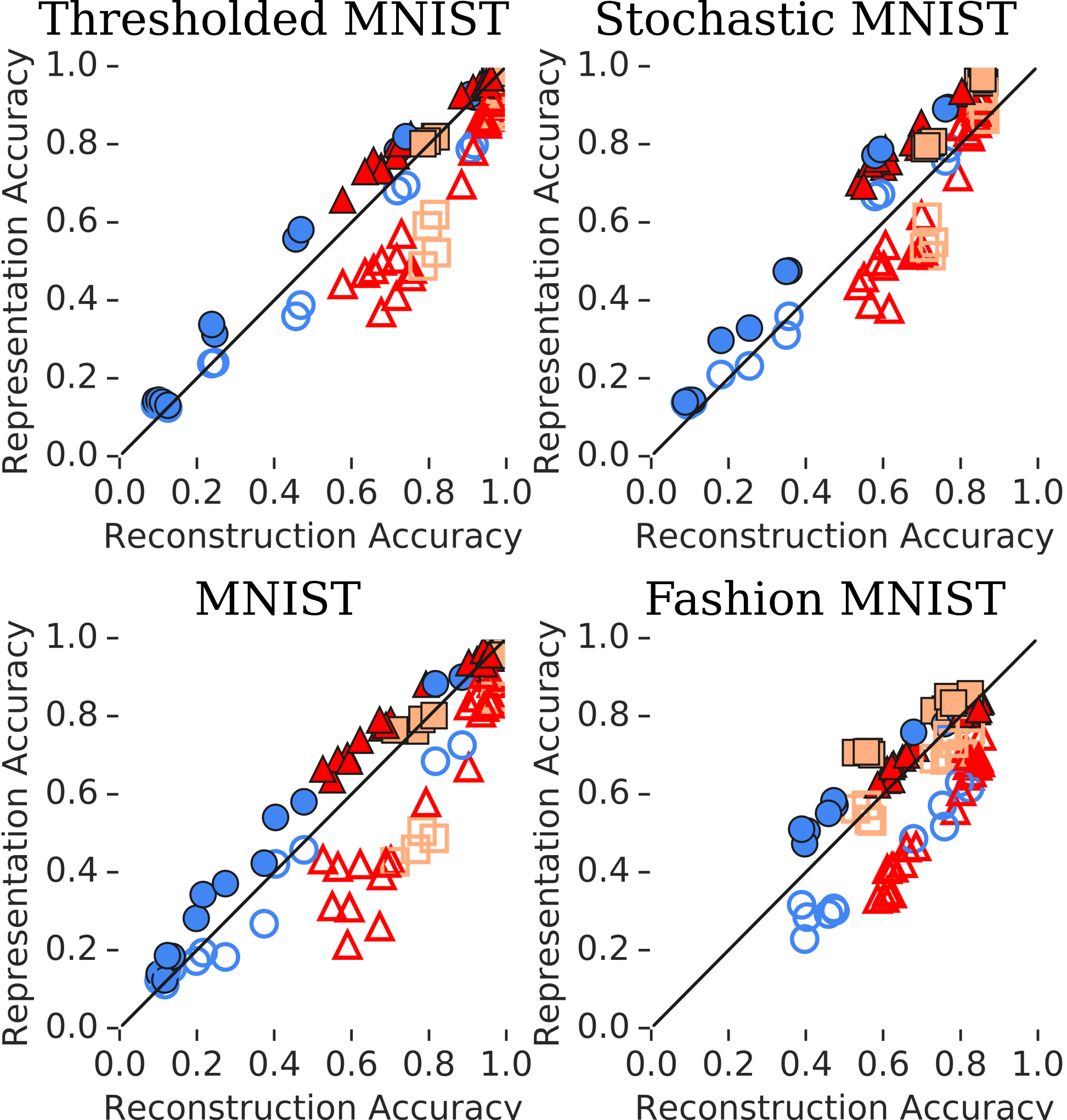}\\
    \end{minipage}
    \vspace{0.1cm}
    \caption{The relationship between the semantic reconstruction accuracy (classifier on $X'$) and the semantic representation accuracy (MLP or linear classifier on $Z$). The MLP is filled and the linear classifier is empty. Symbols and colors mark the decoder types: PixelCNN (\textcolor{blue}{$\bullet$}), 
    CNN (\textcolor[rgb]{1.0, 0.5, 0.0}{$\blacksquare$}), and
    Dueling Decoder (\textcolor{red}{$\blacktriangle$}). Best viewed in color.}
    \label{fig:accuracy_relationships}
    \vspace{-0.5cm}
\end{figure}

The semantic representation accuracy and semantic reconstruction accuracy are different means of estimating the quality of the learned representations. The linear or MLP representation accuracy only incorporates the quality of the encoder, while the reconstruction accuracy depends on the decoder and separate classifier. Even though these are potentially very different, Figure~\ref{fig:accuracy_relationships} shows that all three measures are highly correlated ($r^2>0.8$, Table~\ref{table:accuracy_relationships}). As the horizontal axis is the semantic reconstruction accuracy and the vertical axis also measures accuracy, the line with unity slope is the equivalent performance to the semantic reconstruction accuracy. Most MLP representation classifiers fall above the line for reconstruction accuracy and most linear representation classifiers fall below the line for reconstruction accuracy. The VAE decoder and subsequent classifier can never increase the amount of information about the input (due to the Data Processing Inequality), so it is useful to estimate how much semantic information is lost. While there is some variability across data subsets, on average, the decoder and classifier maintain 88\%$\pm$11\% of the semantic information available to an MLP in the latent space and the linear classifier only maintains 77\%$\pm$15\% of the information. Therefore the semantic reconstruction accuracy is a good proxy for the amount of information the VAE has extracted and also measures the performance of the decoder, which the representation accuracy does not.

\begin{table}%[b!h]
\centering
\setlength\tabcolsep{5pt}
	\begin{tabular}{|c|c|c|}	
	\hline
	Dataset & Linear & MLP \\
	\hline
	Thresholded MNIST & 0.904 & 0.990 \\
	\hline
	Stochastic MNIST & 0.901 & 0.988 \\
	\hline
	MNIST & 0.875 & 0.981 \\
	\hline
	Fashion MNIST &	0.804 & 0.933 \\
    \hline
	\end{tabular}
\vspace{0.1cm}
\caption{The $r^2$ of the relationship between the semantic reconstruction accuracy and the linear or MLP latent decoder accuracy.}
\label{table:accuracy_relationships}
\vspace{-0.1cm}
\end{table}	

\begin{figure}
    \centering
    \begin{minipage}[b]{1.0\columnwidth}
    \centering
    \includegraphics[width=\columnwidth]{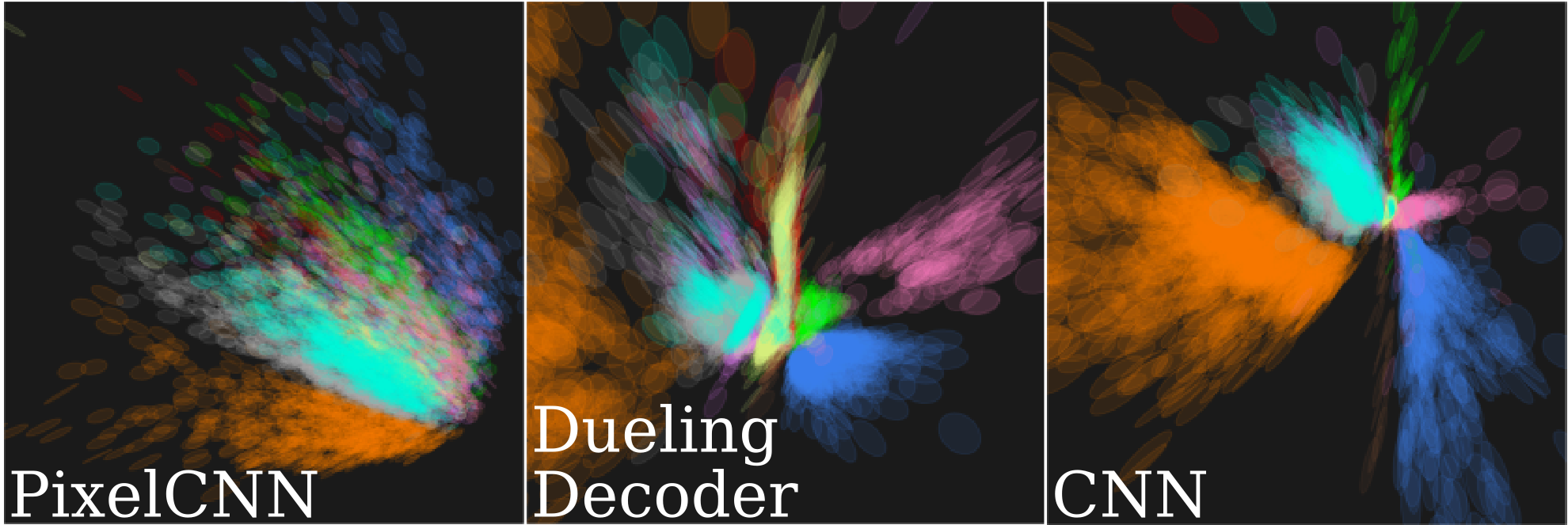}\\
    \end{minipage}
    \caption{Visualization of the 2 dimensional latent space on Thresholded MNIST for the listed decoder models. Each ellipse is located at the mean of each encoder distribution with axes defined by the covariance matrix. The ellipses are color-coded by the label. Best viewed in color.}
    \label{fig:latent_vis}
    \vspace{-0.5cm}
\end{figure}

To understand why different models have different degrees of semantic reconstruction accuracy, we explored the latent spaces of these models. In Figure~\ref{fig:latent_vis}, we visualize the 2D encoder distributions for each test set image and color coded each with the class label. The PixelCNN models exhibit very little structure in the latent space. In contrast, the CNN and Dueling Decoder models exhibit more clustering in the latent space. Similar visualizations for other data sets and t-SNE embedding visulizations of the best low-rate model regardless of latent space dimensionality are available in the supplemental. Each visualization shows a consistent trend: when the model had high semantic reconstruction or representation accuracy, inputs with the same label are visibly clustered together in the latent space.

\section{Conclusions}
We find that autoregressive models are capable of using structure in their latent space when it is present. Adding a pixel-independent auxiliary decoder is an effective way to regularize the latent space for image labeling and helps autoregressive decoders escape the local minima that lead to autodecoders. Regularizing the latent space improves the extraction of semantic information over many other objective manipulations that prevent collapse. The rate, distortion, and ELBO are all poor proxies for semantic information which may explain why optimizing re-weighted ELBO objectives does not generally maximize semantic information. Further work will be needed to understand and explore the right auxiliary tasks for VAEs that maximize semantic information.

{\small
\bibliographystyle{ieee}
\bibliography{egbib}

\begin{thebibliography}{10}\itemsep=-1pt

\bibitem{alemi2018tweet}
A.~Alemi.
\newblock {``Here is a one line MNIST Dataset Loader in Python in a tweet.''},
  2018.
\newblock https://twitter.com/alemi/status/1042834067173957632.

\bibitem{alemi2018fixing}
A.~Alemi, B.~Poole, I.~Fischer, J.~Dillon, R.~A. Saurous, and K.~Murphy.
\newblock Fixing a broken elbo.
\newblock In {\em International Conference on Machine Learning}, pages
  159--168, 2018.

\bibitem{alex2018therml}
A.~A. Alemi and I.~Fischer.
\newblock Therml: Thermodynamics of machine learning, 2018.

\bibitem{alemi2018uncertainty}
A.~A. Alemi, I.~Fischer, and J.~V. Dillon.
\newblock Uncertainty in the variational information bottleneck.
\newblock {\em arXiv preprint arXiv:1807.00906}, 2018.

\bibitem{alemi2016vib}
A.~A. Alemi, I.~Fischer, J.~V. Dillon, and K.~Murphy.
\newblock Deep variational information bottleneck.
\newblock {\em arXiv preprint arXiv:1612.00410}, 2016.

\bibitem{anonymous2019the}
Anonymous.
\newblock The conditional entropy bottleneck.
\newblock In {\em Submitted to International Conference on Learning
  Representations}, 2019.
\newblock under review.

\bibitem{ba2016layer}
J.~L. Ba, J.~R. Kiros, and G.~E. Hinton.
\newblock Layer normalization.
\newblock {\em arXiv preprint arXiv:1607.06450}, 2016.

\bibitem{bengio2013representation}
Y.~Bengio, A.~Courville, and P.~Vincent.
\newblock Representation learning: A review and new perspectives.
\newblock {\em IEEE transactions on pattern analysis and machine intelligence},
  35(8):1798--1828, 2013.

\bibitem{Bishop:2006:PRM:1162264}
C.~M. Bishop.
\newblock {\em Pattern Recognition and Machine Learning (Information Science
  and Statistics)}.
\newblock Springer-Verlag, Berlin, Heidelberg, 2006.

\bibitem{bowman2015generating}
S.~R. Bowman, L.~Vilnis, O.~Vinyals, A.~M. Dai, R.~Jozefowicz, and S.~Bengio.
\newblock Generating sentences from a continuous space, 2015.

\bibitem{burgess2018understanding}
C.~P. Burgess, I.~Higgins, A.~Pal, L.~Matthey, N.~Watters, G.~Desjardins, and
  A.~Lerchner.
\newblock {Understanding disentangling in $beta $-VAE}.
\newblock {\em arXiv preprint arXiv:1804.03599}, 2018.

\bibitem{chen2016lossy}
X.~Chen, D.~P. Kingma, T.~Salimans, Y.~Duan, P.~Dhariwal, J.~Schulman,
  I.~Sutskever, and P.~Abbeel.
\newblock Variational lossy autoencoder.
\newblock {\em arXiv preprint arXiv:1611.02731}, 2016.

\bibitem{clevert2015fast}
D.-A. Clevert, T.~Unterthiner, and S.~Hochreiter.
\newblock Fast and accurate deep network learning by exponential linear units
  (elus), 2015.

\bibitem{davidson2018hyperspherical}
T.~R. Davidson, L.~Falorsi, N.~De~Cao, T.~Kipf, and J.~M. Tomczak.
\newblock Hyperspherical variational auto-encoders.
\newblock {\em arXiv preprint arXiv:1804.00891}, 2018.

\bibitem{dieng2018avoiding}
A.~B. Dieng, Y.~Kim, A.~M. Rush, and D.~M. Blei.
\newblock Avoiding latent variable collapse with generative skip models.
\newblock {\em arXiv preprint arXiv:1807.04863}, 2018.

\bibitem{fertig2018beta}
E.~Fertig, A.~Arbabi, and A.~A. Alemi.
\newblock $\beta$-vaes can retain label information even at high compression.
\newblock {\em Bayesian Deep Learning Workshop at NeurIPS 2018}, 2018.

\bibitem{fischer2014training}
A.~Fischer and C.~Igel.
\newblock Training restricted boltzmann machines: An introduction.
\newblock {\em Pattern Recognition}, 47(1):25--39, 2014.

\bibitem{goodfellow2014generative}
I.~Goodfellow, J.~Pouget-Abadie, M.~Mirza, B.~Xu, D.~Warde-Farley, S.~Ozair,
  A.~Courville, and Y.~Bengio.
\newblock Generative adversarial nets.
\newblock In {\em Advances in neural information processing systems}, pages
  2672--2680, 2014.

\bibitem{gulrajani2016pixelvae}
I.~Gulrajani, K.~Kumar, F.~Ahmed, A.~A. Taiga, F.~Visin, D.~Vazquez, and
  A.~Courville.
\newblock Pixelvae: A latent variable model for natural images.
\newblock {\em arXiv preprint arXiv:1611.05013}, 2016.

\bibitem{hafner2018learning}
D.~{Hafner}, T.~{Lillicrap}, I.~{Fischer}, R.~{Villegas}, D.~{Ha}, H.~{Lee},
  and J.~{Davidson}.
\newblock {Learning Latent Dynamics for Planning from Pixels}.
\newblock {\em ArXiv e-prints}, 2018.

\bibitem{he2019lagging}
J.~He, D.~Spokoyny, G.~Neubig, and T.~Berg-Kirkpatrick.
\newblock Lagging inference networks and posterior collapse in variational
  autoencoders.
\newblock {\em arXiv preprint arXiv:1901.05534}, 2019.

\bibitem{he2016deep}
K.~He, X.~Zhang, S.~Ren, and J.~Sun.
\newblock Deep residual learning for image recognition.
\newblock In {\em Proceedings of the IEEE conference on computer vision and
  pattern recognition}, pages 770--778, 2016.

\bibitem{higgins2017beta}
I.~Higgins, L.~Matthey, A.~Pal, C.~Burgess, X.~Glorot, M.~Botvinick,
  S.~Mohamed, and A.~Lerchner.
\newblock beta-vae: Learning basic visual concepts with a constrained
  variational framework.
\newblock In {\em International Conference on Learning Representations}, 2017.

\bibitem{ioffe2015batch}
S.~Ioffe and C.~Szegedy.
\newblock Batch normalization: Accelerating deep network training by reducing
  internal covariate shift.
\newblock {\em arXiv preprint arXiv:1502.03167}, 2015.

\bibitem{jaderberg2016reinforcement}
M.~Jaderberg, V.~Mnih, W.~M. Czarnecki, T.~Schaul, J.~Z. Leibo, D.~Silver, and
  K.~Kavukcuoglu.
\newblock Reinforcement learning with unsupervised auxiliary tasks.
\newblock {\em arXiv preprint arXiv:1611.05397}, 2016.

\bibitem{kim2018semi}
Y.~Kim, S.~Wiseman, A.~Miller, D.~Sontag, and A.~Rush.
\newblock Semi-amortized variational autoencoders.
\newblock In {\em International Conference on Machine Learning}, pages
  2683--2692, 2018.

\bibitem{kingma2014semi}
D.~P. Kingma, S.~Mohamed, D.~Jimenez~Rezende, and M.~Welling.
\newblock Semi-supervised learning with deep generative models.
\newblock In Z.~Ghahramani, M.~Welling, C.~Cortes, N.~D. Lawrence, and K.~Q.
  Weinberger, editors, {\em Advances in Neural Information Processing Systems
  27}, pages 3581--3589. Curran Associates, Inc., 2014.

\bibitem{kingma2016improving}
D.~P. Kingma, T.~Salimans, R.~Jozefowicz, X.~Chen, I.~Sutskever, and
  M.~Welling.
\newblock Improving variational inference with inverse autoregressive flow,
  2016.

\bibitem{kingma2013auto}
D.~P. Kingma and M.~Welling.
\newblock Auto-encoding variational bayes.
\newblock {\em arXiv preprint arXiv:1312.6114}, 2013.

\bibitem{lecun2010mnist}
Y.~LeCun, C.~Cortes, and C.~Burges.
\newblock Mnist handwritten digit database.
\newblock {\em AT\&T Labs [Online]. Available: http://yann. lecun.
  com/exdb/mnist}, 2, 2010.

\bibitem{lucas2018auxiliary}
T.~Lucas and J.~Verbeek.
\newblock Auxiliary guided autoregressive variational autoencoders, 2018.

\bibitem{Murphy:2012:MLP:2380985}
K.~P. Murphy.
\newblock {\em Machine Learning: A Probabilistic Perspective}.
\newblock The MIT Press, 2012.

\bibitem{oord2016pixel}
A.~v.~d. Oord, N.~Kalchbrenner, and K.~Kavukcuoglu.
\newblock Pixel recurrent neural networks.
\newblock {\em arXiv preprint arXiv:1601.06759}, 2016.

\bibitem{salakhutdinov2008quantitative}
R.~Salakhutdinov and I.~Murray.
\newblock On the quantitative analysis of deep belief networks.
\newblock In {\em Proceedings of the 25th international conference on Machine
  learning}, pages 872--879. ACM, 2008.

\bibitem{salimans2017pixelcnn}
T.~Salimans, A.~Karpathy, X.~Chen, and D.~P. Kingma.
\newblock Pixelcnn++: Improving the pixelcnn with discretized logistic mixture
  likelihood and other modifications.
\newblock {\em arXiv preprint arXiv:1701.05517}, 2017.

\bibitem{Spurr_2018_CVPR}
A.~Spurr, J.~Song, S.~Park, and O.~Hilliges.
\newblock Cross-modal deep variational hand pose estimation.
\newblock In {\em The IEEE Conference on Computer Vision and Pattern
  Recognition (CVPR)}, June 2018.

\bibitem{srivastava2014dropout}
N.~Srivastava, G.~Hinton, A.~Krizhevsky, I.~Sutskever, and R.~Salakhutdinov.
\newblock Dropout: a simple way to prevent neural networks from overfitting.
\newblock {\em The Journal of Machine Learning Research}, 15(1):1929--1958,
  2014.

\bibitem{szegedy2015going}
C.~Szegedy, W.~Liu, Y.~Jia, P.~Sermanet, S.~Reed, D.~Anguelov, D.~Erhan,
  V.~Vanhoucke, and A.~Rabinovich.
\newblock Going deeper with convolutions.
\newblock In {\em Proceedings of the IEEE conference on computer vision and
  pattern recognition}, pages 1--9, 2015.

\bibitem{tomczak2017vae}
J.~M. Tomczak and M.~Welling.
\newblock Vae with a vampprior.
\newblock {\em arXiv preprint arXiv:1705.07120}, 2017.

\bibitem{van2016conditional}
A.~van~den Oord, N.~Kalchbrenner, L.~Espeholt, O.~Vinyals, A.~Graves, et~al.
\newblock Conditional image generation with pixelcnn decoders.
\newblock In {\em Advances in Neural Information Processing Systems}, pages
  4790--4798, 2016.

\bibitem{van2017neural}
A.~van~den Oord, O.~Vinyals, et~al.
\newblock Neural discrete representation learning.
\newblock In {\em Advances in Neural Information Processing Systems}, pages
  6306--6315, 2017.

\bibitem{vedantam2017generative}
R.~Vedantam, I.~Fischer, J.~Huang, and K.~Murphy.
\newblock Generative models of visually grounded imagination.
\newblock {\em arXiv preprint arXiv:1705.10762}, 2017.

\bibitem{wu2018group}
Y.~Wu and K.~He.
\newblock Group normalization.
\newblock {\em arXiv preprint arXiv:1803.08494}, 2018.

\bibitem{xiao2017fashion}
H.~Xiao, K.~Rasul, and R.~Vollgraf.
\newblock Fashion-mnist: a novel image dataset for benchmarking machine
  learning algorithms.
\newblock {\em arXiv preprint arXiv:1708.07747}, 2017.

\bibitem{yang2017improved}
Z.~Yang, Z.~Hu, R.~Salakhutdinov, and T.~Berg-Kirkpatrick.
\newblock Improved variational autoencoders for text modeling using dilated
  convolutions.
\newblock In {\em Proceedings of the 34th International Conference on Machine
  Learning-Volume 70}, pages 3881--3890. JMLR. org, 2017.

\bibitem{zagoruyko2016wide}
S.~Zagoruyko and N.~Komodakis.
\newblock Wide residual networks, 2016.

\bibitem{Zhao2017}
S.~Zhao, J.~Song, and S.~Ermon.
\newblock {InfoVAE: Information Maximizing Variational Autoencoders}.
\newblock jun 2017.

\bibitem{zhao2017learning}
T.~Zhao, R.~Zhao, and M.~Eskenazi.
\newblock Learning discourse-level diversity for neural dialog models using
  conditional variational autoencoders.
\newblock {\em Proceedings of the 55th Annual Meeting of the Association for
  Computational Linguistics}, 2017.

\end{thebibliography}
}

\clearpage

\section*{Supplemental}

\subsection{Network Architecture Details}

\begin{table*}%[b!h]
\centering
\setlength\tabcolsep{5pt}
	\begin{tabular}{|c|c|c|c|c|c|}	
	\hline
	\shortstack{Encoder\\ layer} & Filters & Shape & Activation & Stride & Padding \\
	\hline
	1 & 32 & [5, 5] & relu & 1 & same \\
	\hline
	2 & 64 & [5, 5] & relu & 2 & same \\
	\hline
	3 & 64 & [5, 5] & relu & 1 & same \\
	\hline
	4 & 64 & [5, 5] & relu & 2 & same \\
	\hline
	5 & 128 & [7, 7] & relu & 1 & valid \\
	\hline
	6 & \shortstack{distribution\\ parameters} & FC & None & & \\
    \hline
	\end{tabular}
\vspace{0.1cm}
\caption{Encoder Architecture. The network for the convolutional encoder. The first 5 layers are all convolutional layers. The last layer is a fully-connected (FC) layer to output the mean and lower-triangular covariance of the Gaussian distribution for the latent size of the current model.}
\label{table:encoder_architecture}
\vspace{-0.1cm}
\end{table*}

\begin{table*}%[b!h]
\centering
\setlength\tabcolsep{5pt}
	\begin{tabular}{|c|c|c|c|c|c|}	
	\hline
	\shortstack{Decoder\\ layer} & Filters & Shape & Activation & Stride & Padding \\
	\hline
	1 & 128 & [7, 7] & relu & 1 & valid \\
	\hline
	2 & 128 & [5, 5] & relu & 1 & same \\
	\hline
	3 & 128 & [5, 5] & relu & 2 & same \\
	\hline
	4 & 64 & [5, 5] & relu & 1 & same \\
	\hline
	5 & 64 & [5, 5] & relu & 2 & same \\
	\hline
	6 & 32 & [5, 5] & relu & 1 & same \\
	\hline
	7 & \shortstack{distribution\\ parameters} & [5, 5] & None & 1 & same \\
    \hline
	\end{tabular}
\vspace{0.1cm}
\caption{Decoder Architecture. The network for the transposed convolution decoder. The first 5 layers are all transposed convolutional layers. The last layer outputs the probability for the Bernoulli distribution or 5 quantized logistic parameters (center, slope, confidence) for each pixel.}
\label{table:decoder_architecture}
\vspace{-0.1cm}
\end{table*}

\begin{table*}%[b!h]
\centering
\setlength\tabcolsep{5pt}
	\begin{tabular}{|c|c|c|c|c|c|}	
	\hline
	\shortstack{PixelCNN\\ block} & Filters & Shape & Activation & Stride & Type \\
	\hline
	1 & 64 / 128 & [2, \{3, 2\}] & sigmoid * tanh & 1 & Down, Right \\
    \hline
	2 & 64 / 128 & [2, \{3, 2\}] & sigmoid * tanh & 1 & Down, Right \\
    \hline
	3 & 64 / 128 & [2, \{3, 2\}] & sigmoid * tanh & 2 & Down, Right \\
    \hline
	4 & 64 / 128 & [2, \{3, 2\}] & sigmoid * tanh & 1 & Down, Right \\
    \hline
	  & - / 128 & [2, \{3, 2\}] & sigmoid * tanh & 1 & Down, Right \\
    \hline
	  & - / 128 & [2, \{3, 2\}] & sigmoid * tanh & 1 & Down, Right \\
    \hline
	5 & 64 / 128 & [2, \{3, 2\}] & sigmoid * tanh & 2 & Down, Right \\
    \hline
	6 & 64 / 128 & [2, \{3, 2\}] & sigmoid * tanh & 1 & Down, Right \\
    \hline
	  & - / 128 & [2, \{3, 2\}] & sigmoid * tanh & 1 & Down, Right \\
    \hline
	  & - / 128 & [2, \{3, 2\}] & sigmoid * tanh & 1 & Down, Right \\
    \hline
	7 & 64 / 128 & [2, \{3, 2\}] & sigmoid * tanh & 2 & Deconv: Down, Right  \\
    \hline
	8 & 64 / 128 & [2, \{3, 2\}] & sigmoid * tanh & 1 & Down, Right \\
    \hline
	  & - / 128 & [2, \{3, 2\}] & sigmoid * tanh & 1 & Down, Right \\
    \hline
	  & - / 128 & [2, \{3, 2\}] & sigmoid * tanh & 1 & Down, Right \\
    \hline
	9 & 64 / 128 & [2, \{3, 2\}] & sigmoid * tanh & 2 & Deconv: Down, Right \\
    \hline
	10 & 64 / 128 & [2, \{3, 2\}] & sigmoid * tanh & 1 & Down, Right \\
    \hline
	  & - / 128 & [2, \{3, 2\}] & sigmoid * tanh & 1 & Down, Right \\
    \hline
	  & - / 128 & [2, \{3, 2\}] & sigmoid * tanh & 1 & Down, Right \\
    \hline
    11 & \shortstack{distribution\\ parameters} & [1, 1] & None & 1 & Conv \\
    \hline
	\end{tabular}
\vspace{0.1cm}
\caption{PixelCNN Architecture. The PixelCNN++ network architecture~\cite{salimans2017pixelcnn}. Each line is block of down and a down-right masked convolution (implemented via shifts). Where the type is listed as ``Deconv'', the transposed convolution with stride 2 increases the dimensions. Skip connections are applied as in the paper between the lower and upper halves of the network. The encoded latent vector is added as a bias to each layer. The output distribution parameters follow the conventions of Table~\ref{table:decoder_architecture}. The un-numbered blocks and larger number of filters are the sizes for the enlarged PixelCNN used in a subset of Fashion MNIST experiments.}
\label{table:pixelcnn_architecture}
\vspace{-0.1cm}
\end{table*}

All models used in this paper are convolutional models unless specified as otherwise. The architecture for the encoder models is listed in Supplemental Table~\ref{table:encoder_architecture}. The architecture for the pixel-independent convolutional decoder models is listed in Supplemental Table~\ref{table:decoder_architecture}. The architecture for the PixelCNN decoder is based off of~\cite{salimans2017pixelcnn} but slightly modified as in Supplemental Table~\ref{table:pixelcnn_architecture}. The auxiliary decoders for intensity histograms or marginal histograms were two fully connected layers of 256 units and ReLU activations feeding into the distribution. The classifier architecture for MNIST based models is the same as the encoder architecture in Supplemental Table~\ref{table:encoder_architecture} but the distribution parameters are replaced with a 10-way softmax classification head. The classifier for the more challenging Fashion MNIST data set is the WideResnet model of~\cite{zagoruyko2016wide}. The linear semantic representation decoder is a single layer, fully-connected network connected to a softmax classifier. The MLP semantic representation decoder has two layers of 200 fully-connected units with Exponential Linear Unit activations followed by a softmax classifier~\cite{clevert2015fast}.

\subsection{Hyperparameters for Best Low-Rate Models}

\begin{table*}%[b!h]
\centering
\setlength\tabcolsep{5pt}
	\begin{tabular}{|c|c|c|c|c||c|c|c|c|}
\hline
Decoder & ELBO & Distortion & Rate & SR Accuracy & $\beta$ & batch size & latent size & regularization \\
\hline
\multicolumn{9}{|c|}{\textbf{THRESHOLDED MNIST}} \\
\hline
PixelCNN & 60.6 & 52.6 $\pm$  0.2 & 8.01 $\pm$ 0.18 & 0.47 $\pm$ 0.07 & 0.1 & 64 & 2 & 0.0 \\
\hline
CNN & 130.4 & 123.4 $\pm$  0.3 & 7.04 $\pm$ 0.11 & 0.81 $\pm$ 0.02 & 1.0 & 64 & 2 & 0.0 \\
\hline
Dueling Decoder & 56.4 & 48.4 $\pm$  0.2 & 7.93 $\pm$ 0.18 & 0.88 $\pm$ 0.01 & 1.0 & 32 & 16 & 0.1 \\
\hline
\multicolumn{9}{|c|}{\textbf{STOCHASTIC MNIST}} \\
\hline
PixelCNN & 83.2 & 74.6 $\pm$  0.1 & 8.57 $\pm$ 0.16 & 0.36 $\pm$ 0.03 & 0.1 & 32 & 2 & 0.0 \\
\hline
CNN & 136.7 & 129.8 $\pm$  0.3 & 6.91 $\pm$ 0.09 & 0.73 $\pm$ 0.02 & 1.0 & 64 & 2 & 0.0 \\
\hline
Dueling Decoder & 81.3 & 72.1 $\pm$  0.8 & 9.25 $\pm$ 0.72 & 0.79 $\pm$ 0.03 & 1.0 & 32 & 16 & 0.1 \\
\hline
\multicolumn{9}{|c|}{\textbf{MNIST}} \\
\hline
PixelCNN & 560.1 & 552.8 $\pm$  4.5 & 7.29 $\pm$ 0.08 & 0.37 $\pm$ 0.02 & 0.1 & 64 & 2 & 0.0 \\
\hline
CNN & 891.6 & 883.6 $\pm$  73.2 & 7.94 $\pm$ 0.28 & 0.78 $\pm$ 0.01 & 1.0 & 64 & 2 & 0.0 \\
\hline
Dueling Decoder & 564.6 & 554.8 $\pm$  2.7 & 9.78 $\pm$ 0.96 & 0.79 $\pm$ 0.04 & 1.0 & 32 & 16 & 0.1 \\
\hline
\multicolumn{9}{|c|}{\textbf{FASHION MNIST}} \\
\hline
PixelCNN & 1492.2 & 1484.0 $\pm$  1.7 & 8.24 $\pm$ 0.17 & 0.76 $\pm$ 0.02 & 1.0 & 64 & 16 & 0.0 \\
\hline
CNN & 1995.3 & 1986.0 $\pm$  1.3 & 9.31 $\pm$ 0.04 & 0.56 $\pm$ 0.00 & 1.0 & 64 & 2 & 0.0 \\
\hline
Dueling Decoder & 1495.1 & 1486.4 $\pm$  2.0 & 8.76 $\pm$ 1.34 & 0.69 $\pm$ 0.03 & 1.0 & 64 & 64 & 0.1 \\
\hline
\hline
PixelCNN+ & 1473.3 & 1469.2 $\pm$  3.2 & 4.08 $\pm$ 0.81 & 0.43 $\pm$ 0.12 & 1.0 & 32 & 16 & 0.0 \\
\hline
Dueling Decoder+ & 1480.2 & 1471.2 $\pm$  4.0 & 9.03 $\pm$ 0.11 & 0.67 $\pm$ 0.01 & 1.0 & 32 & 2 & 1.0 \\
\hline
\end{tabular}
\vspace{0.2cm}
\caption{For each data set and each decoder, we report results for the model with rate$<$10 and the highest semantic representation accuracy. The reported ELBO is the sum of the distortion and rate. The distortion, rate, and reconstruction accuracy are the means of three randomly initialized training runs $\pm$ one standard deviation. The hyperparameters, $\beta$, batch size, number of latent dimensions, and degree of Dueling Decoder regularization, $\lambda$, used to train these models are on the right. The decoders with + are enlarged PixelCNN decoders described in Table~\ref{table:pixelcnn_architecture}}
\label{table:supp_best_results}
\end{table*}

The hyperparameters for training the best models with rates$<$10 for each data set are provided in Supplemental Table~\ref{table:supp_best_results} in the same order as in the main text.

\subsection{Distortion-Rate Trade-Off}
As in~\cite{alemi2018fixing}, across the range of models tested models the ELBO is bounded by the limited information content of the data set, but models can trade off between rate and distortion. In Supplemental Figure~\ref{fig:supp_rate_distortion}, the models are plotted in the rate-distortion plane and color coded for semantic reconstruction accuracy. As shown in the main text, the accuracy increases rapidly at low rates and is often stable as the rate increases further. For the MNIST and Fashion MNIST data sets, we do not plot the CNN models in Supplemental Figure~\ref{fig:supp_rate_distortion} because their distortions are hundreds of nats worse and visualizing them obscures the rest of the data.

\begin{figure}[b]
    \centering
    \begin{minipage}[b]{0.99\columnwidth}
    \includegraphics[width=\columnwidth]{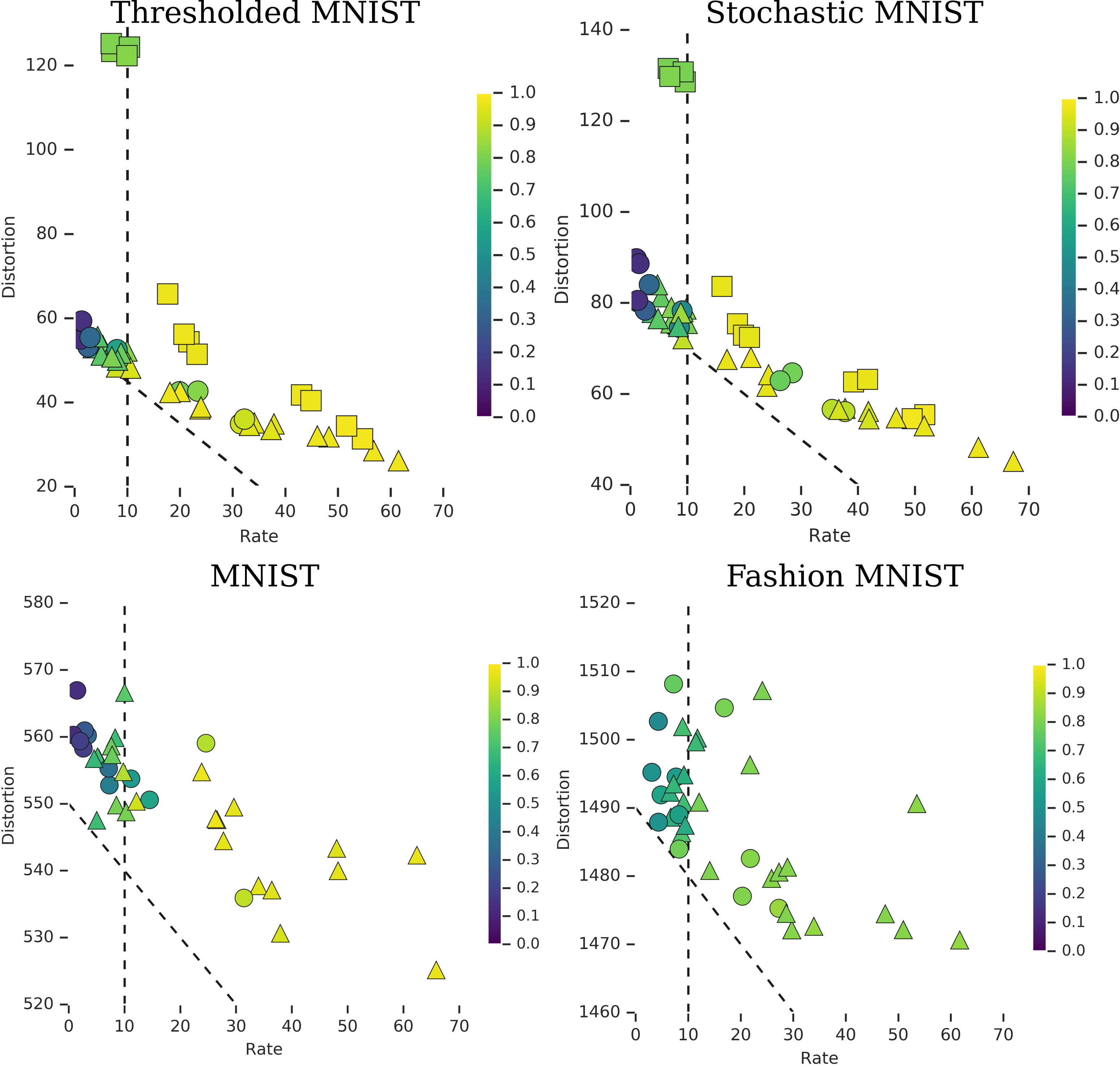}\\
    \end{minipage}
    \vspace{-0.2cm}
    \caption{Rate, distortion, and accuracy for models on each dataset. Shape indicates the architecture: PixelCNN ($\bullet$), CNN ($\blacksquare$), and Dueling Decoder ($\blacktriangle$). Color represents the semantic reconstruction accuracy. The dashed line at a rate of 10 nats marks the cutoff for the low-rate models. The diagonal line marks a constant ELBO contour. For the MNIST and Fashion MNIST data sets, the CNN decoder models have much higher distortions than are visible in this plot and are not represented here.}
    \label{fig:supp_rate_distortion}
    \vspace{-0.5cm}
\end{figure}

\subsection{Latent Space Visualizations}

\begin{figure*}
    \centering
    \begin{minipage}[b]{0.8\textwidth}
    \includegraphics[width=\textwidth]{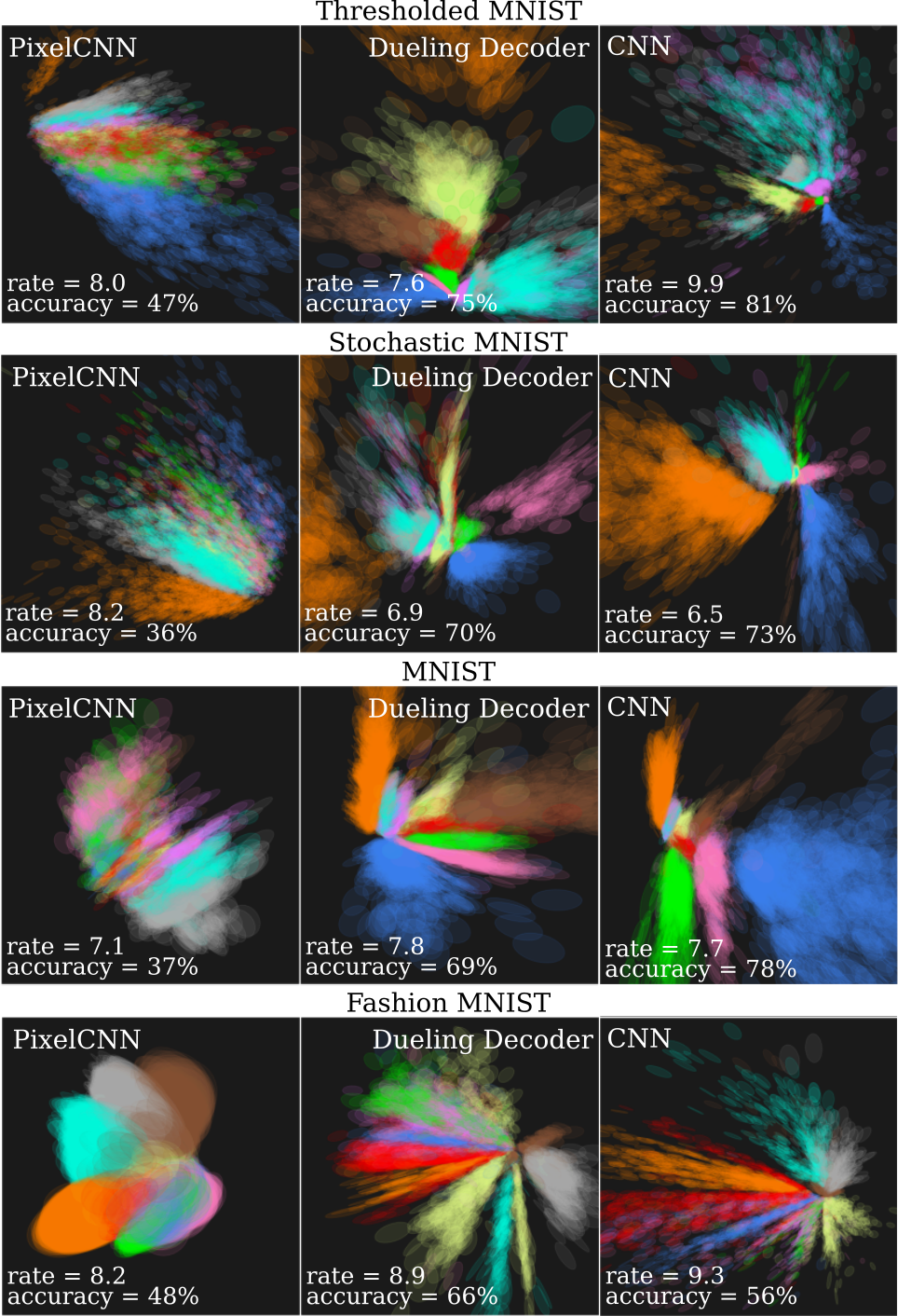}\\
    \end{minipage}
    \vspace{-0.2cm}
    \caption{Visualization of the 2 dimensional latent space on each data set for most accurate low-rate decoder models with 2 latent dimensions. Each ellipse is located at the mean of each encoder distribution with axes defined by the covariance matrix. The ellipses are color-coded by the label. The rate and semantic reconstruction accuracy are listed for each model. Best viewed in color.}
    \label{fig:supp_latent2_vis}
    \vspace{-0.5cm}
\end{figure*}

\begin{figure*}
    \centering
    \begin{minipage}[b]{0.8\textwidth}
    \includegraphics[width=\textwidth]{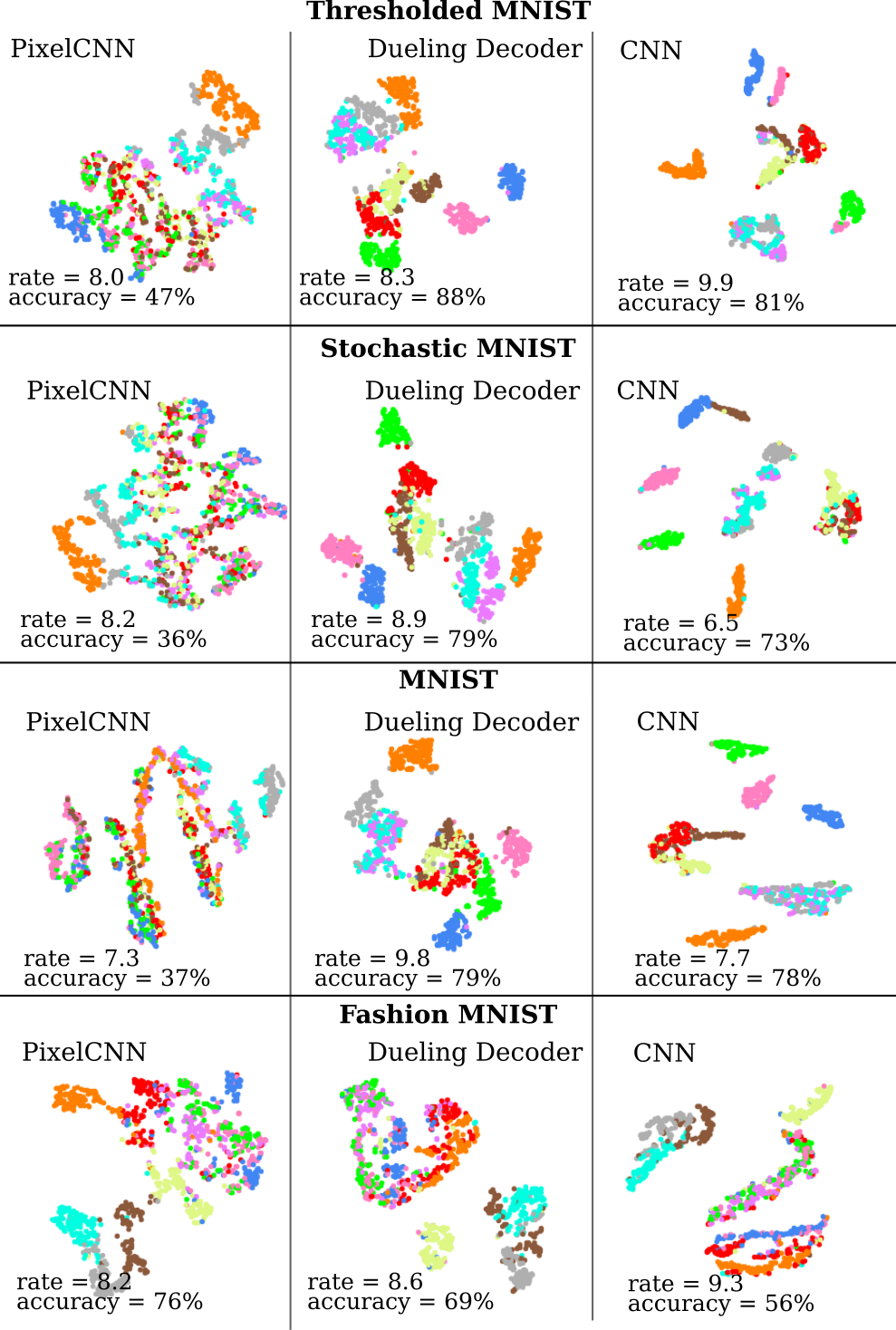}\\
    \end{minipage}
    \vspace{-0.2cm}
    \caption{Visualization of the 2 dimensional t-SNE of the latent latent space on each data set for most accurate low-rate decoder models with any number of latent dimensions. Each point is the t-SNE value for the mean of the encoder distribution on the test set, color-coded by the label. The rate and semantic reconstruction accuracy are listed for each model. Best viewed in color.}
    \label{fig:supp_latent_tsne}
    \vspace{-0.5cm}
\end{figure*}

\begin{table}
\centering
\setlength\tabcolsep{5pt}
\begin{tabular}{|c|c|c|}
\hline
Color & MNIST & Fashion MNIST \\
\hline
{\color[RGB]{66, 134, 244}\huge$\bullet$} & 0 & T-shirt/top \\
\hline
{\color[RGB]{255, 128, 0}\huge$\bullet$} & 1 & Trouser \\
\hline
{\color[RGB]{0, 255, 0}\huge$\bullet$} & 2 & Pullover \\
\hline
{\color[RGB]{255, 0, 0}\huge$\bullet$} & 3 & Dress \\
\hline
{\color[RGB]{235, 122, 255}\huge$\bullet$} & 4 & Coat \\
\hline
{\color[RGB]{140, 88, 58}\huge$\bullet$} & 5 & Sandal \\
\hline
{\color[RGB]{255, 127, 193}\huge$\bullet$} & 6 & Shirt \\
\hline
{\color[RGB]{175, 175, 175}\huge$\bullet$} & 7 & Sneaker \\
\hline
{\color[RGB]{222, 247, 133}\huge$\bullet$} & 8 & Bag \\
\hline
{\color[RGB]{0, 255, 225}\huge$\bullet$} & 9 & Ankle boot \\
\hline
	\end{tabular}
\vspace{0.2cm}
\caption{The mapping between color and data set labels used in the latent space visualizations. Best viewed in color.}
\label{table:supp_color_code}
\end{table}

In Supplemental Figure~\ref{fig:supp_latent2_vis}, we visualize the 2 dimensional encoder distributions for each test set image and color coded each with the class label for each data set. We can only directly visualize the best models with 2 dimensional latent spaces, so we use t-SNE for the best low-rate models regardless of latent space dimensionality in Supplemental Figure~\ref{fig:supp_latent_tsne}. As reported in the main text, each visualization shows a consistent trend: when the model had high semantic reconstruction accuracy, inputs with the same label are visibly clustered together in the latent space. The color code for each label is provided in Supplemental Table~\ref{table:supp_color_code}.

\end{document}